\documentclass[runningheads]{llncs}
\usepackage[T1]{fontenc}
\usepackage[misc]{ifsym}
\usepackage{cite}
\usepackage{amsmath,amssymb,amsfonts}
\usepackage{graphicx}
\usepackage{textcomp}
\usepackage{xcolor}
\usepackage[ruled,vlined]{algorithm2e}
\usepackage{booktabs} 
\usepackage{multirow} 
\usepackage{array}    
\usepackage{tabularx} 
\usepackage{pifont} 
\usepackage{booktabs}
\usepackage{adjustbox}
\usepackage{paralist}
\usepackage{bbding}

\newcommand{\lzw}[1]{{\textcolor{black}{#1}}}
\newcommand{\zwc}[1]{{\textcolor{black}{#1}}}
\newcommand{\zwcforshorter}[1]{{\textcolor{black}{#1}}}

\begin{document}
\title{Multi-Representation Adapter with Neural Architecture Search for Efficient Range-Doppler Radar Object Detection}

\author{Zhiwei Lin$^{1}$\thanks{Equal contribution. $\dagger$ Corresponding author. \\This work was done when Weicheng was an intern at PKU.}, Weicheng Zheng$^{2 \star}$, Yongtao Wang$^{1\dagger}$}
\institute{$^{1}$Wangxuan Institute of Computer Technology, Peking University, China.\\
$^{2}$ School of Computer Science and Technology, Tongji University, China.\\
\{zwlin, wyt\}@pku.edu.cn\\
}
\maketitle              
\begin{abstract}
Detecting objects efficiently from radar sensors has recently become a popular trend due to their robustness against adverse lighting and weather conditions compared with cameras. This paper presents an efficient object detection model for Range-Doppler (RD) radar maps. Specifically, we first represent RD radar maps with multi-representation, \textit{i.e.}, heatmaps and grayscale images, to gather high-level object and fine-grained texture features. Then, we design an additional Adapter branch, an Exchanger Module with two modes, and a Primary-Auxiliary Fusion Module to effectively extract, exchange, and fuse features from the multi-representation inputs, respectively. Furthermore, we construct a supernet with various width and fusion operations in the Adapter branch for the proposed model and employ a One-Shot Neural Architecture Search method to further improve the model's efficiency while maintaining high performance. Experimental results demonstrate that our model obtains favorable accuracy and efficiency trade-off. Moreover, we achieve new state-of-the-art performance on RADDet and CARRADA datasets with mAP@50 of 71.9 and 57.1, respectively.

\end{abstract}
\section{Introduction}
\label{sec:intro}

Object detection is a fundamental task in computer vision, aiming to locate and \lzw{classify} objects within \lzw{various inputs, including images and radar}, and has found widespread applications across various domains, \lzw{such as autonomous driving and robots.}
\lzw{This capability enables intelligent agents to perceive the surrounding environments and adjust their behaviors.
Typically, detection models use images as the input.
However, images from camera sensors are often sensitive to lighting and weather conditions, resulting in a low robust detection performance.}

\lzw{Recently, radar sensors have exhibited lower dependency on lighting and weather conditions~\cite{Qian2021Robust} and can provide 3D information on the distance and velocity of objects.}
Specifically, distance information facilitates the prediction of objects’ relative positions, while velocity is instrumental in analyzing objects' motion. 
More recently, publicly available radar datasets such as RADDet~\cite{RADDet9469418}, and CARRADA~\cite{CARRADA9413181} have significantly advanced research in radar-based object detection.

To utilize raw radar data, which contains more information than point clouds format, current methods usually transform it into Range-Doppler (RD)~\cite{DAROD9827281,lssiclrod}, Range-Angle (RA)~\cite{Zheng2021Scene-aware}, and Range-Angle-Doppler (RAD) maps~\cite{BoostRad}.
Among them, the RD map is usually a very efficient representation in the radar signal processing chain to detect objects~\cite{lssiclrod}.
Nevertheless, \lzw{current RD detection} methods often fail to fully leverage the multi-representation potential of RD maps, particularly the integration of complementary features like intensity and spatial information. 
Moreover, they \lzw{lack the design of} lightweight networks, leaving significant room for improvement in balancing accuracy and efficiency to meet the computational constraints of edge devices.

\lzw{To address this issue, we} propose an efficient radar object detection model, which is built upon a YOLO-type framework~\cite{yolov8}, augmented with \lzw{multiple representations and an Adapter branch to enhance feature extraction.}
\lzw{Specifically, we leverage the multi-representation capabilities of RD maps, \textit{i.e.,} heatmaps and grayscale images, as inputs.}
The heatmaps contain rich contextual information and are fed into the YOLO-type backbone branch, while grayscale images preserving finer-grained and raw details are directed to the Adapter branch.
In addition, to facilitate \lzw{multi-representation feature extraction}, we introduce a Primary-Auxiliary Fusion Module for feature fusion, alongside \zwc{an Exchanger Module with two modes that enables} bidirectional information exchange between the Backbone and Adapter branches.
This design significantly enhances the model’s capability to extract and utilize features effectively from the multi-representation of RD maps. \lzw{Moreover, to accommodate resource-constrained real-world scenarios, we further introduce Neural Architecture Search (NAS) by forming a weight-sharing supernet with adaptive widths and fusion strategies in the Adapter branch.}
The main contributions of this work are:
\begin{compactitem}

    \item \lzw{We propose a novel paradigm for enhancing feature extraction in RD maps by adopting a multi-representation of RD map and integrating an Adapter branch into the YOLO-type backbone branch with bidirectional feature interaction between multi-representation features.}

    \item \lzw{We design a Primary-Auxiliary Fusion Module for multi-representation feature fusion and \zwc{an Exchanger Module with two symmetric modes} to enable seamless feature exchange between the backbone and Adapter branches.}

    \item \lzw{We employ a One-Shot NAS framework to search for the best adapter module and further improve the efficiency of the proposed model.}

    \item \lzw{Comprehensive experiments on RADDet and CARRADA datasets demonstrate the proposed model achieves state-of-the-art detection performance, as well as the best accuracy and efficient trade-off.}
    
\end{compactitem}

\section{Related Works}

\subsection{Efficient Object Detectors}

Object detection methods are broadly categorized into two classes: two-stage and single-stage approaches. 
Two-stage methods, such as Faster R-CNN~\cite{faster7485869}, generate region proposals \lzw{in the first stage and adopt prediction heads in the second stage to refine the proposal bounding boxes}, offering higher accuracy but slower inference. 
Single-stage methods, such as YOLO~\cite{yolov1} and SSD~\cite{SSD}, directly predict bounding boxes and classes \lzw{from predefined anchors or generated objectness heatmaps, achieving superior accuracy and efficiency trade-off and more favored by practical applications.}

Among single-stage detectors, the YOLO series is the most representative. 
Specifically, YOLO~\cite{yolov1} \lzw{first proposes to divide an image into regular grid cells and predicts bounding boxes in each cell,} enabling detection in a single network pass.
From YOLOv2~\cite{yolov2} to YOLOv5~\cite{yolov5}, \lzw{they define a set of anchors for each grid cell in YOLO and introduce several improvements, including high-resolution and efficient backbones, complex training strategies, and multi-scale predictions.}
YOLOv6~\cite{yolov6} and YOLOX~\cite{ge2021yolox} mark the key advancement by designing an anchor-free pipeline \lzw{and decoupling the regression and classification heads.}
YOLOv8~\cite{yolov8}, renowned for its strong performance and extensibility, has become the foundation for subsequent iterations, including YOLOv9-12~\cite{yolov9,yolov10,yolov11,yolov12}.

In this paper, we adopt the YOLO-type detector as the baseline due to its versatility, efficiency, and compatibility, and introduce the Adapter branch and NAS to obtain models with a better accuracy and efficiency trade-off. 

\subsection{Object Detection on RD Maps} 
\lzw{In recent years, radar object detection has gained increasing attention. 
Among different representations of radar data, the RD map is a very efficient representation for detecting objects.}
However, compared to RAD maps, studies focusing exclusively on RD maps remain limited.
DAROD~\cite{DAROD9827281} introduces a Faster R-CNN-based framework for RD maps, incorporating Doppler velocity into feature vectors to enhance detection performance, but its simple convolutional design limits the extraction of multi-level features. 
\lzw{RADDet~\cite{RADDet9469418} designs a single network to process RAD and RD maps.}  
However, its ResNet-based backbone and dual detection heads lead to high computational complexity and poor performance on RD maps.
RiCL~\cite{lssiclrod} introduces semi-supervised learning for RD map by enhancing FCOS with \lzw{the radar instance contrastive learning.}
Though it achieves competitive accuracy with limited labeled data, RiCL relies on a complex pretraining pipeline and struggles to extract features directly from RD maps.

Despite these advances, existing methods often fail to fully utilize the multi-representation capabilities of RD maps, such as combining heatmaps and grayscale images for collaborative feature extraction. 
To address these challenges, we propose \lzw{incorporating} an Adapter branch to enhance the accuracy and efficiency of RD map feature extraction.

\subsection{Neural Architecture Search} 
NAS aims to automate the design of lightweight and high-performance network architectures with constraints.
\lzw{
Early NAS methods directly enumerate or randomly sample and train numerous architectures from scratch. Thus, their computational cost is high, and they are only suitable for small datasets.}
To improve search efficiency, the Evolutionary Algorithm is introduced into NAS~\cite{zhou2020EcoNAS}, reducing the search complexity by selectively exploring candidate architectures.
Furthermore, to address the high training costs of a large set of subnets, the One-Shot and \lzw{Zero-shot} NAS method is proposed~\cite{guo2020single,liu2019DARTS,bnnas} \lzw{by training a supernet that contains all potential subnets within the search space.}
\lzw{
These methods sample subnets using weights from the trained supernet and evaluate with a predefined performance metric on a validation set during the search phase.}

In this paper, we apply One-Shot NAS, \lzw{which is more flexible than Zero-shot NAS}, to the proposed model to further optimize the accuracy and efficiency trade-off.

\begin{figure*}[t!]
\centering
\includegraphics[width=\textwidth]{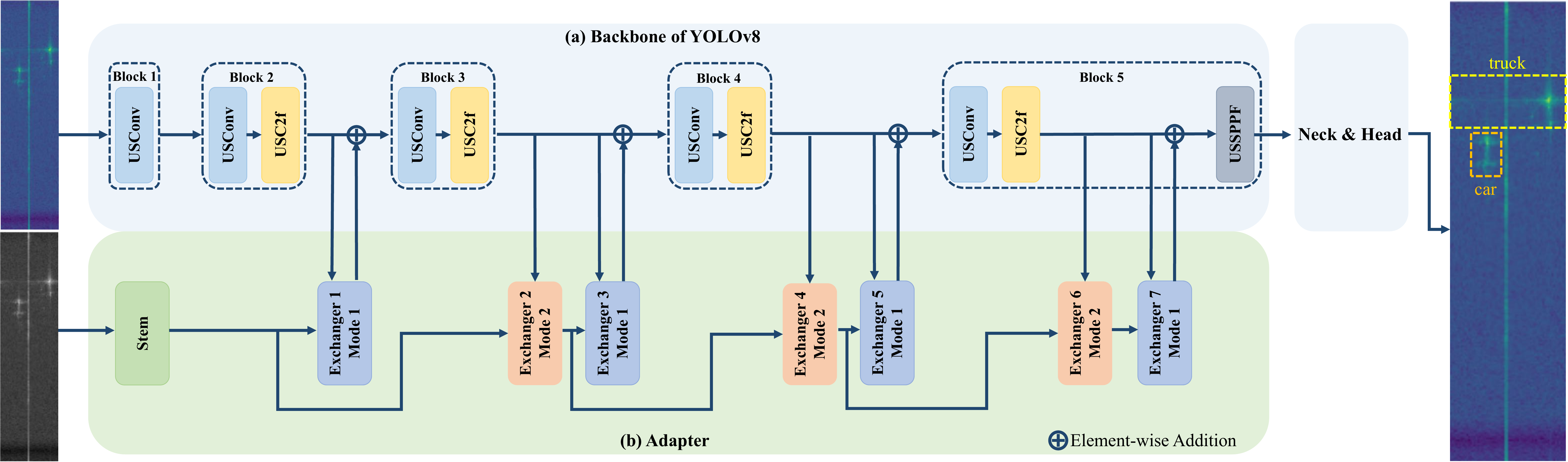}

\caption{\textbf{The \lzw{overall pipeline} of YOLOv8n-Adapter.} (a) The searchable YOLOv8 backbone branch takes the heatmap representation of RD maps as inputs. (b) The proposed Adapter branch uses the grayscale representation of RD maps as inputs.
Exchanger Modules, alternating between two modes, establish a symmetric information flow, \lzw{enhancing the overall feature extraction} ability from RD maps.
}
\label{arc}
\vspace{-10pt}
\end{figure*}

\section{METHODOLOGY}

\subsection{Overall Architecture}\label{AA}
Without loss of generality, we take YOLOv8 as the baseline detector for illustration.
As shown in Fig.\ref{arc}, the proposed model comprises two branches, \textit{i.e.}, the searchable YOLOv8 backbone branch and the Adapter branch, to extract and fuse features from multi-representation RD maps.
Specifically, as shown in Fig.\ref{arc} (a), the searchable YOLOv8 backbone branch takes the heatmap representation of the RD map as input and consists of a series of feature extraction blocks.
The first two blocks primarily extract low-level features from the heatmap input. 
Then, starting from the third block, each block incorporates features from the Adapter branch into the current heatmap feature map $X_{\textit{heat}}$, enriching its feature representation. 
For the Adapter branch, as depicted in Fig.\ref{arc} (b), it takes the grayscale representation of the RD map as input and establishes multi-level feature interactions with the backbone branch. 
Specifically, the grayscale representation input first passes through the Stem module for preliminary feature extraction and scale adjustment, which is a simple feature extraction module consisting of three convolutional layers.
\lzw{Then, in each Exchanger Module following, the grayscale feature maps $X_{\textit{gray}}$ interact with the heatmap feature maps $X_{\textit{heat}}$ from the backbone branch.}
The fused features are routed to different branches depending on the mode of the Exchanger Module.
\lzw{
After feature extraction from two branches, we send the features to the searchable YOLOv8 neck and head to obtain the final object detection predictions.
}

\subsection{Multi-Representation Input}\label{sec:multi_rep_input}

\lzw{Our model adopts} the multi-representation by feeding two different representations of the same RD map, \textit{i.e.}, the heatmap and grayscale image, into the backbone and Adapter \lzw{branch}, respectively.
Specifically, the heatmap representation of the RD map is created using pseudo color encoding to highlight different value ranges, enhancing visual contrast and making objects with distinct velocity or distance characteristics more prominent. 
This aids the backbone branch in capturing 
\lzw{high-level object} features and motion patterns efficiently. 
In contrast, the grayscale representation retains the original numerical distribution from the RD map, providing more intensity details to reveal edges and textures and support low-level feature extraction. 
%
\lzw{In addition, we specially design the Adapter branch to further capture fine-grained details from the grayscale representation and complement the high-level features extracted by the backbone branch.}

\begin{figure}[t!]
\centering
\includegraphics[width=0.6\textwidth]{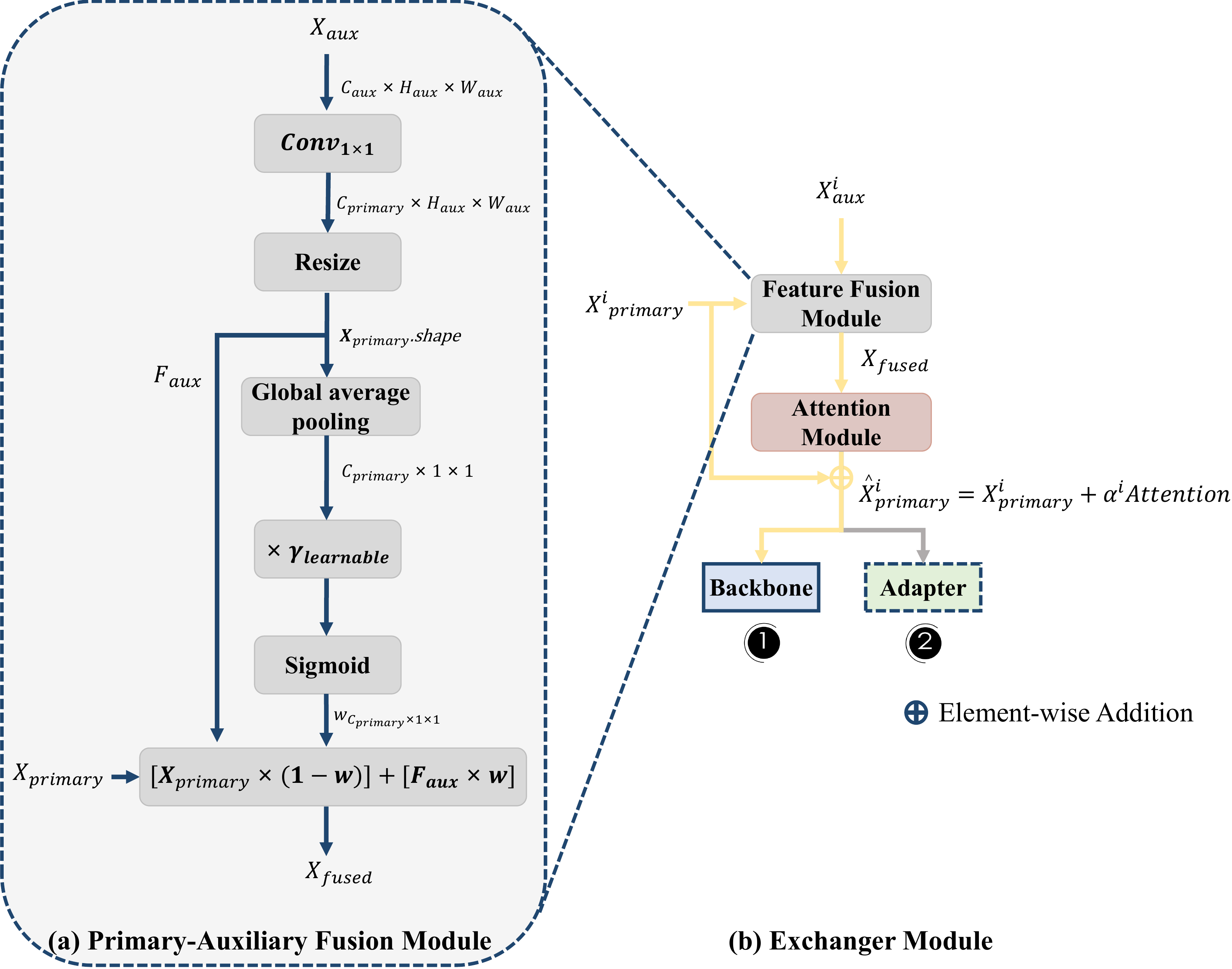}
\caption{\textbf{The key components of the Adapter branch.} (a) Primary-Auxiliary Fusion Module is designed to integrate primary and auxiliary feature maps.
(b) Exchanger Module will exchange features from two branches.
}

\label{adapterModules}
\vspace{-10pt}
\end{figure}

\subsection{Adapter}\label{sec:adapter}
\lzw{As shown in Fig.\ref{arc} (b), the Adapter consists of a Stem module for preliminary feature extraction and several Exchanger Modules, which alternate between two modes, establishing a symmetric information exchange with the backbone branch.}
Each Exchanger Module includes the key processes of primary-auxiliary feature fusion, attention computation, and residual connections.

\noindent\textbf{Primary-Auxiliary Fusion Module.}
\label{Primary-Auxiliary Fusion Module}
The Primary-Auxiliary Fusion module is an essential component in each Exchanger Module. 
\lzw{It is designed to integrate two feature maps, \textit{i.e.}, primary feature and auxiliary feature from two branches, by preserving the core information from the primary feature and enhancing it with complementary information from the auxiliary feature.}
\zwcforshorter{Specifically, as shown in Fig.\ref{adapterModules} (a), the auxiliary feature map undergoes a 1 $\times$ 1 convolution to match the primary feature map's channel count and is resized to align spatial dimensions, resulting in $F_{\textit{aux}}$. Channel-level features from $F_{\textit{aux}}$ are extracted via global average pooling, scaled by a learnable parameter $\gamma$ with a Sigmoid activation, generating a weighting factor $w$. $X_{\textit{primary}}$ and $F_{\textit{aux}}$ are then fused using a weighted sum with $w$, producing the fused feature map $X_{\textit{fused}}$.}


\noindent\textbf{Exchanger Module.}
The Exchanger Module is designed to facilitate bidirectional feature interaction between backbone and Adapter branches, operating in two distinct modes. In mode 1, the feature map $X_{\textit{heat}}$ from the backbone branch serves as the primary feature map, while the feature map $X_{\textit{gray}}$ from the Adapter branch acts as the auxiliary feature map. This configuration is intended to inject low-level, detailed grayscale features into the $X_{\textit{heat}}$ within the Backbone branch. In mode 2, the roles of the primary and auxiliary feature maps are reversed.
Specifically, we first send the primary and auxiliary feature maps, $X_{\textit{primary}}$ and $X_{\textit{aux}}$, to the Primary-Auxiliary Fusion Module to obtain a fused feature map $X_{\textit{fused}}$. Then, $X_{\textit{fused}}$ is fed into an attention module, where attention weights are applied to $X_{\textit{fused}}$. Finally, $X_{\textit{primary}}$ and the output of the attention module are combined through a weighted residual connection as follows:
\begin{align}
\hat{X}_{\textit{primary}}^{i} &= X_{\textit{primary}}^{i}+\alpha^i\text{Attention}(X_{\textit{fused}}^{i}), \label{eq1}
\end{align}
where $i$ denotes the ID of the Exchanger Module, and the $\text{Attention}(\cdot)$ is a lightweight Coordinate Attention \cite{Hou_2021_CoordinateAttention}, and $\alpha^i$ is a learnable balance weights.

\begin{figure}[t]
    \centering
    \begin{minipage}{0.52\linewidth}
        \centering
    \includegraphics[width=\linewidth]{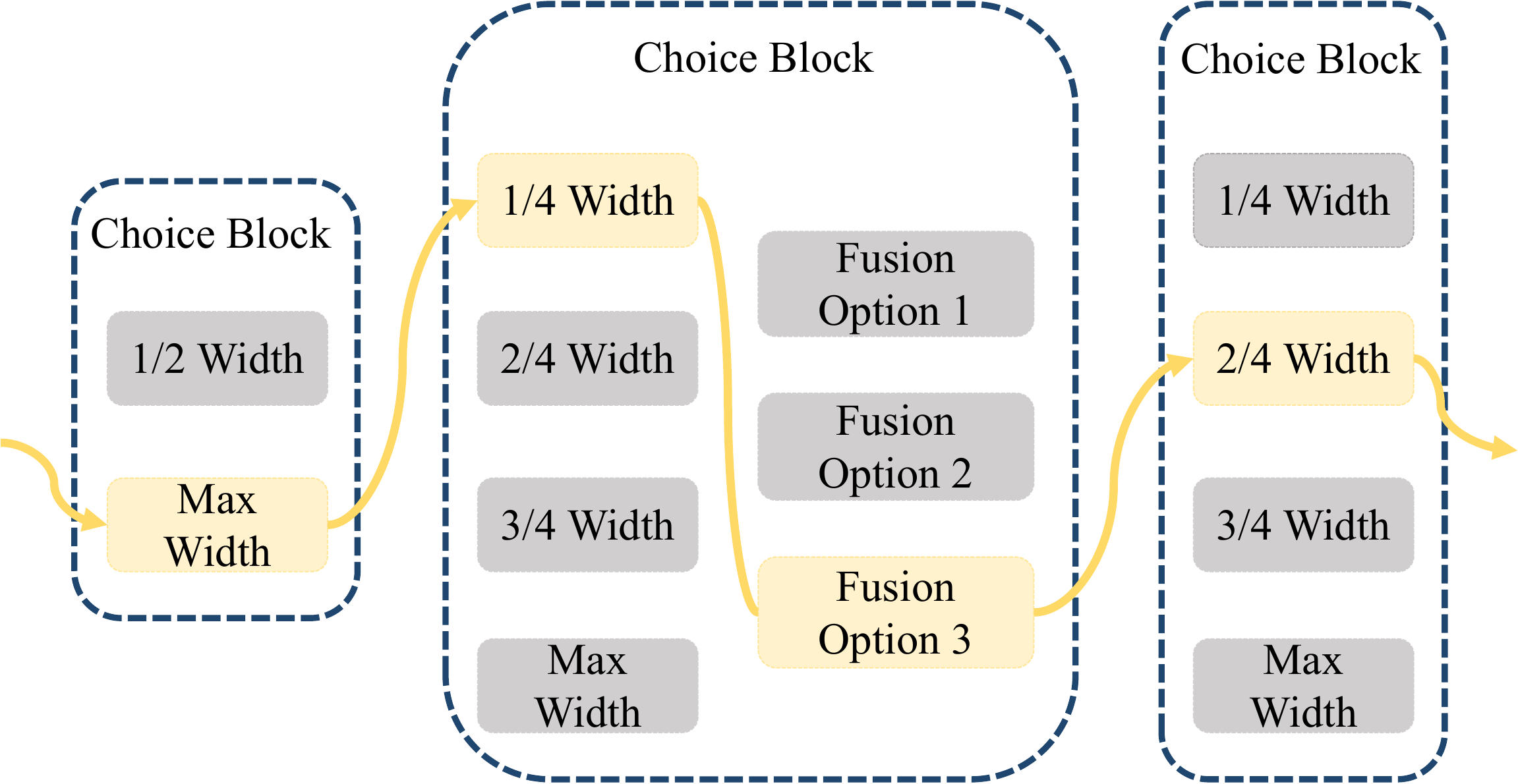}
        \caption{
         \textbf{Illustration of weight-sharing supernet.} 
        During training and inference, one option of each Choice Block is activated.
        }
        \label{single_path_supernet}
    \end{minipage}
    \hfill 
    \begin{minipage}{0.45\linewidth}
        \centering
    \includegraphics[width=\linewidth]{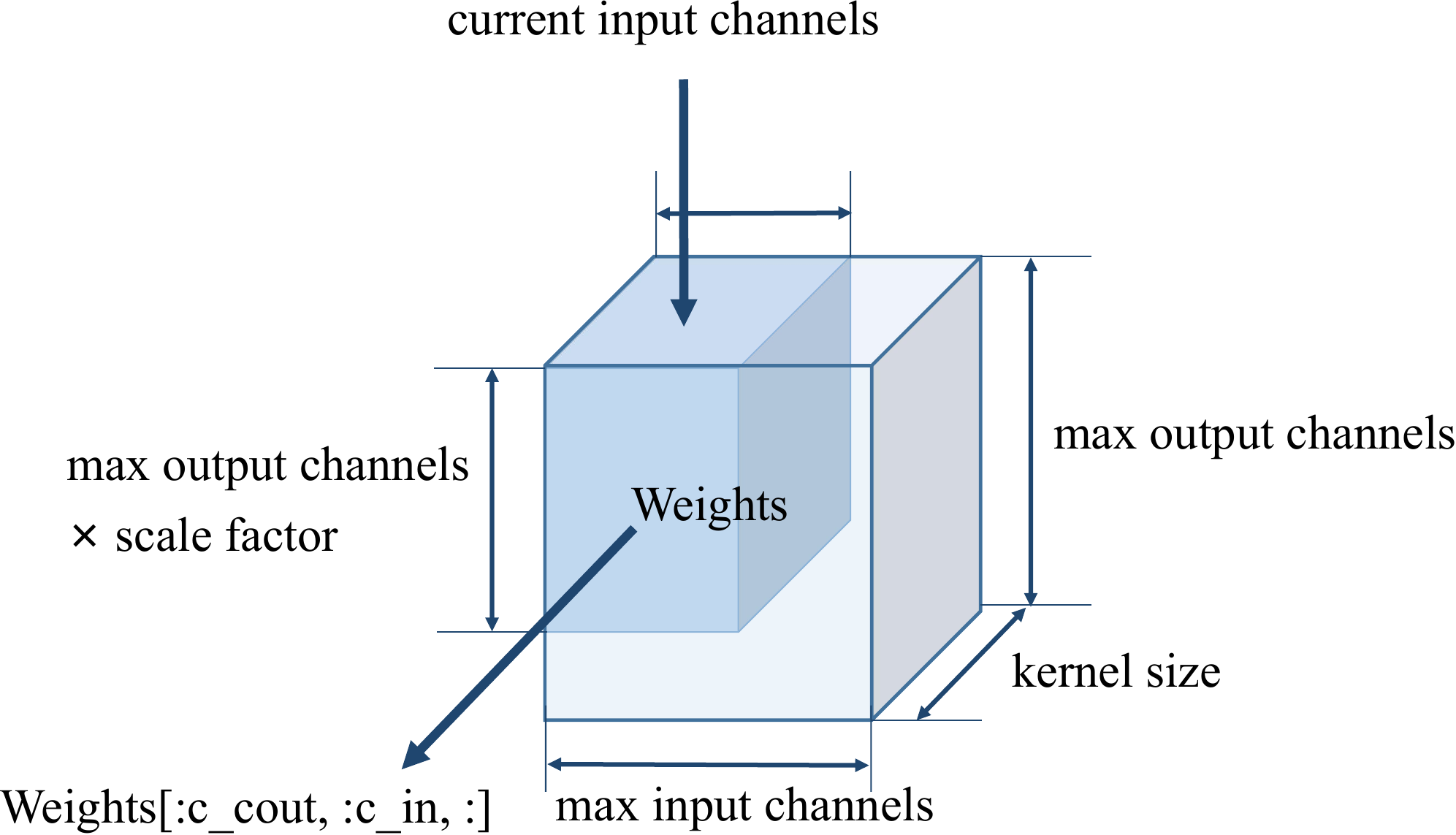}
        \caption{\textbf{Elastic convolution kernel based on weight sharing.} \lzw{We utilize dynamic channel pruning to obtain the selected weights.}
        }
        \label{dynamic_channel_block}
    \end{minipage}
    \vspace{-10pt}
\end{figure}

\begin{table}[t]
\centering
\caption{\textbf{The overview of search space.} \lzw{Each component in our model is searchable.}}
\resizebox{0.8\columnwidth}{!}{
\begin{tabular}{l|c|c|c|c}
\toprule
\textbf{} & \textbf{Component} & \textbf{Width Options} & \textbf{Fusion Options}& \textbf{Search Space Size} \\
\midrule
\multirow{2}{*}{Backbone} & Block 1-2 & (1/2, 1)&- & \multirow{6}{*}{$2^{20} \times 3^7$} \\
         & Block 3-5 & (1/4, 1/2, 3/4, 1)& -&  \\
\cmidrule{1-4}
\multirow{2}{*}{Adapter} & Stem \zwcforshorter{(3 blocks)} & (1/4, 1/2, 3/4, 1)&- &  \\

& Exchanger 1-7 & \multirow{1}{*}{Adaptive} & (1, 2, 3)& \\
\cmidrule{1-4}
Neck \& Head & Block 1-3 & (1/4, 1/2, 3/4, 1)& -&  \\
\bottomrule
\end{tabular}
}
\label{tab:search_space}
\vspace{-10pt}
\end{table}

\subsection{One-Shot NAS}\label{sec:NAS}
\noindent\textbf{Supernet Construction with Elastic Width and Variable Feature Fusion.}
Given the search space $\mathcal{A}$, the supernet is represented as $\mathcal{N}(\mathcal{A}, W)$ with a shared weight $W$ for all subnets.
The training of the supernet can be formulated as follows:
\vspace{-5pt}
\begin{align}
W_{\mathcal{A}} = \arg\min_W \mathcal{L}_{\text{train}}(\mathcal{N}(\mathcal{A}, W))
,\label{supernet_training}
\vspace{-7pt}
\end{align}
where $\mathcal{L}_{\text{train}}(\cdot)$ is the loss function on the training set.

For our model, the constructed supernet consists of multiple Choice Blocks, as shown in Fig.~\ref{single_path_supernet}. 
Each Choice Blocks are implemented based on a weight-sharing mechanism with elastic width.
In a Choice Block like Exchanger, the operation for fusing primary and auxiliary features is designed to be variable, providing multiple fusion options.
During each forward pass in training, only one sampled configuration within each Choice Block is activated.

Specifically, for elastic width, a scaling factor is randomly sampled from the Choice Block to determine the current convolution width.
Then, weight tensors corresponding to the specific input and output widths are dynamically pruned from the largest kernel weight tensor for forward computation, as illustrated in Fig. \ref{dynamic_channel_block}.
Based on this strategy, the standard convolution module in YOLOv8 is modified to create the Uniform Sampling Convolution (USConv), enabling dynamic weight tensor pruning.

\noindent\textbf{Search Space $\mathcal{A}$.}
The supernet is constructed based on the YOLOv8n-Adapter. Specifically, as shown in Table \ref{tab:search_space}, we list the width options and the feature fusion options in the search phase.
\lzw{All convolution layers can be searched to adjust their convolution width dynamically.}
Exchanger modules adjust their width to align with backbone and Stem outputs.
Within each Exchanger, the Searchable Primary-Auxiliary Fusion Module supports three feature fusion methods. Beyond the approach proposed in Section~\ref{Primary-Auxiliary Fusion Module} (serving as Option 1), two simpler fusion methods are introduced: element-wise summation (Option 2) and weighted summation with a learnable coefficient (Option 3).
\noindent\textbf{Evolutionary Architecture Search.}
The goal of the search phase can be described as follows:
\vspace{-5pt}
\begin{equation}
a^{*} = \arg\max_{a \in \mathcal{A}} \text{ACC}_{\text{val}} ( \mathcal{N}(a, W_{\mathcal{A}}(a)) ),\label{subnet_searching}
\end{equation}
\zwcforshorter{where $W_{\mathcal{A}}(a)$ represents weights inherited by subnet $a$ from the trained supernet, and $\mathrm{ACC}_{\text{val}}(\mathcal{N}(a, W_{\mathcal{A}}(a)))$ denotes the validation accuracy achieved by subnet $a$ after performing forward inference with these weights, which requires very low computational cost.}
\zwcforshorter{However, the vast search space with billions of subnet architectures makes exhaustive evaluation resource-intensive, while random sampling often yields sub-optimal results.}
To address these challenges, we employ an evolutionary algorithm to \lzw{search for the best subnet efficiently.}
Subnets are randomly sampled initially to achieve population size $P$, and subsequently maintained through Crossover and Mutation operations.
During each iteration, each candidate subnet $a$ is ranked based on its accuracy, and only the top $k$ subnets are preserved for the next round.
The Crossover operation involves randomly selecting two subnets from the candidates and generating a new subnet by randomly choosing the architecture configuration for each Choice Block from one of the two selected subnets.
Mutation denotes randomly selecting a subnet from the candidates and applying a probability \textit{prob} to modify the architecture choice of each Choice Block, creating a new subnet.  
Moreover, constraints such as model parameters or computational cost can be applied, retaining only subnets that satisfy them as candidates.
\section{Experiments}

\subsection{Experiments Settings}
We evaluate the proposed model on two public radar datasets, RADDet \cite{RADDet9469418} and CARRADA \cite{CARRADA9413181}. 
For the RADDet dataset, we extract the raw analog-to-digital-converter data and process it using range and Doppler FFT to generate RD spectrograms. 
For the CARRADA dataset, we directly adopt the RD spectrograms from the official data.

We adopt YOLOv8 as our baseline model. The model is trained for 300 epochs with a batch size of 64, the SGD optimizer, and an initial learning rate of 0.01. 
The learnable parameters $\gamma$ in the Primary-Auxiliary Fusion Module and $\alpha$ in the Exchanger Module are initialized to zero. 
For the evolutionary architecture search, the population size $P$ is 50, the maximum iterations $\mathcal{T}$ is 20, and the top $k = 15$ subnets are retained per iteration. After the search, the top 5 subnets are fully trained, and the best-performing one is selected.

During inference, we set the IoU threshold for Non-Maximum Suppression to 0.1 following previous works~\cite{DAROD9827281}. We adopt mean Average Precision (mAP) as the primary metric for evaluation.
The inference time is estimated on a CPU (Intel(R) Xeon(R) Gold 6248R).

\begin{table}[ht]
\centering
\begin{minipage}{0.48\textwidth}
\centering
\caption{\textbf{Comparison of object detection results on the RADDet and CARRADA datasets.} 
The best results are shown in bold, and the second-best results are underlined.}
\resizebox{\columnwidth}{!}{
\begin{tabular}{l|l|c|c|c}
\toprule
\textbf{Dataset} & \textbf{Model} & \textbf{Params (M)} & \textbf{mAP@30} & \textbf{mAP@50} \\
\midrule
\multirow{6}{*}{\textbf{RADDet}} 
 & DAROD~\cite{DAROD9827281}                       & \underline{3.46}  & \underline{65.6} & \underline{46.6} \\
 & RADDet RD~\cite{RADDet9469418}                   & 7.89  & 38.4          & 22.9 \\
 & FCOS~\cite{lssiclrod}                                 & -  & -             & 39.7 \\
 & FCOS + RiCL~\cite{lssiclrod}                          & -  & -             & 41.0 \\
\cmidrule(lr){2-5}
 & \textbf{YOLOv8n-Adapter-Search}  & \textbf{2.65} & \textbf{75.4} & \textbf{71.9} \\
\midrule
\multirow{6}{*}{\textbf{CARRADA}} 
 & DAROD \cite{DAROD9827281}                       & \underline{3.46}  & \textbf{70.7} & \underline{55.8} \\
 & RADDet RD~\cite{RADDet9469418}                   & 7.89  & 48.6          & 18.6 \\
 & FCOS~\cite{lssiclrod}                                 & -  & -             & 44.5 \\
 & FCOS + RiCL~\cite{lssiclrod}                          & -  & -             & 49.2 \\
\cmidrule(lr){2-5}
 & \textbf{YOLOv8n-Adapter-Search}  & \textbf{2.50} & \underline{62.1} & \textbf{57.1} \\
\bottomrule
\end{tabular}}
\label{SOTA}
\end{minipage}
\hfill
\begin{minipage}{0.48\textwidth}
\centering
\caption{\textbf{Comparison of object detection results of the searched subnet and YOLOv8 Variants.} 
The best results are shown in bold, and the second-best results are underlined.}
\vspace{-3pt}
\resizebox{\columnwidth}{!}{%
\begin{tabular}{l|l|c|c|c}
\toprule
\textbf{Dataset} & \textbf{Model} & \textbf{Params (M)} & \textbf{mAP@50} & \textbf{mAP@50-95} \\
\midrule
\multirow{6}{*}{\textbf{RADDet}} 
 & YOLOv8n                      & 3.01  & 65.0  & 42.2    \\
 & YOLOv8s                      & 11.13 & 67.6  & 42.6    \\
 & YOLOv8m                      & 25.84 & 67.9  & 43.2    \\
 & YOLOv8l                      & 43.63 & \underline{70.8}  & 43.2 \\
 & YOLOv8n-Adapter              & 3.24  & 69.8  & \underline{45.1}  \\
\cmidrule(lr){2-5}
 & \textbf{YOLOv8n-Adapter-Search}  & \textbf{2.65} & \textbf{71.9}  & \textbf{45.4}   \\
\midrule
\multirow{6}{*}{\textbf{CARRADA}} 
 & YOLOv8n                      & 3.01  & 55.5  & 28.0  \\
 & YOLOv8s                      & 11.13 & \underline{56.7}  & 28.6  \\
 & YOLOv8m                      & 25.84 & 54.4  & 26.1 \\
 & YOLOv8l                      & 43.63 & 52.6  & 24.0  \\
 & YOLOv8n-Adapter              & 3.24  & 56.6  & \underline{29.6}  \\
\cmidrule(lr){2-5}
 & \textbf{YOLOv8n-Adapter-Search}  & \textbf{2.50} & \textbf{57.1}  & \textbf{30.3}  \\
\bottomrule
\end{tabular}%
}
\label{searchResultTable}
\end{minipage}
\vspace{-3pt}
\end{table}

\subsection{Main Results}
Table \ref{SOTA} presents a detailed comparison of the best subnet with the state-of-the-art methods on the RADDet and CARRADA datasets.
The results show that our method achieves new state-of-the-art performance compared with previous methods.
Specifically, on the RADDet dataset, our best subnet uses only 2.65M parameters while \lzw{supassing DAROD and FCOS + RiCL by 25.3 and 30.9 mAP@50, respectively.}
\lzw{For the CARRADA dataset, our model achieves higher mAP@50, outperforming DAROD by 1.3 mAP@50.}

Our best subnet demonstrates favorable detection performance on multiple key metrics while significantly reducing model parameters. 
This highlights the proposed approach's potential to achieve efficient radar object detection under constrained computational resources.

\begin{figure*}[t]
    \centering
    \begin{minipage}{0.32\linewidth}
        \centering
        \includegraphics[width=\linewidth]{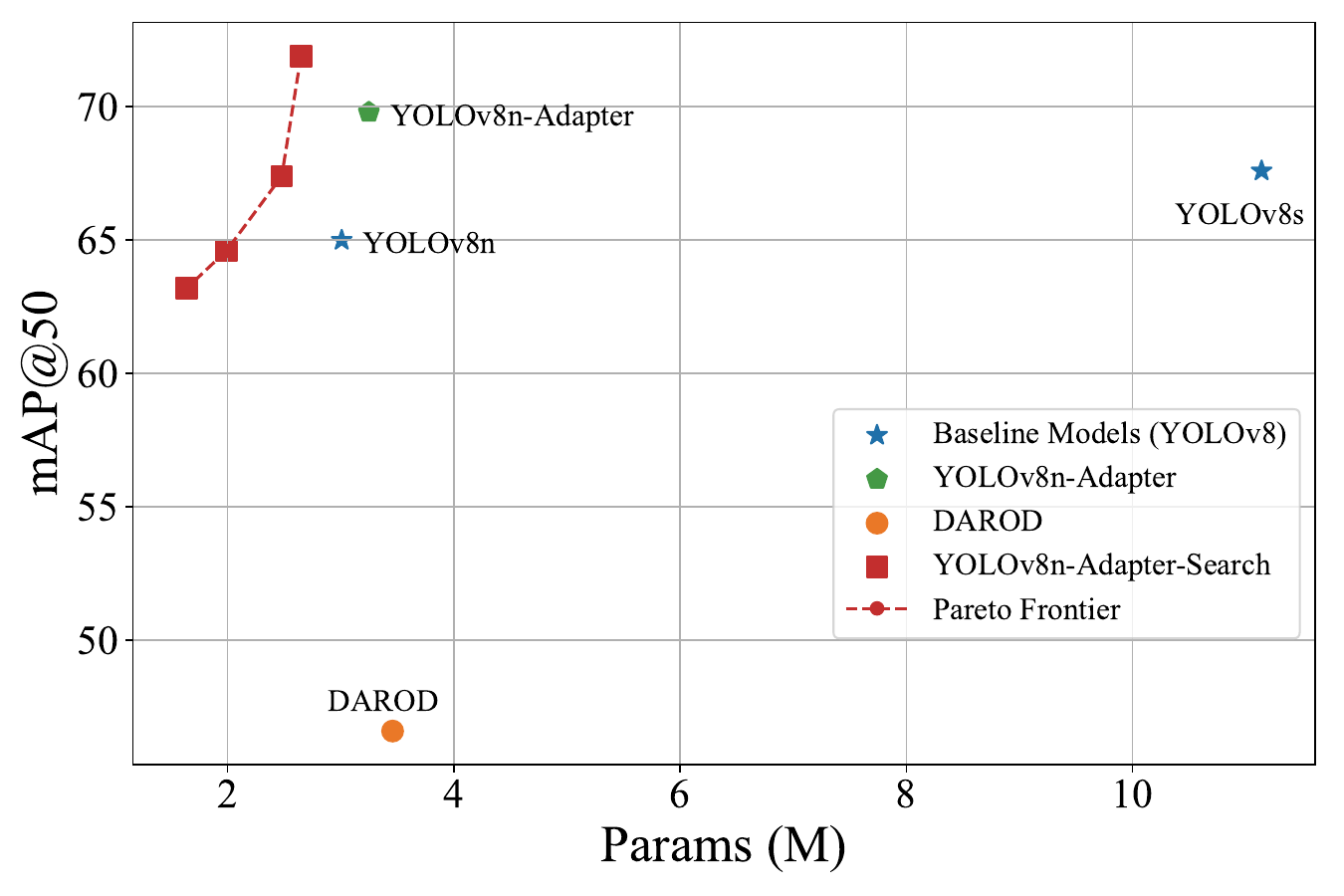}
    \end{minipage}
    \begin{minipage}{0.32\linewidth}
        \centering
        \includegraphics[width=\linewidth]{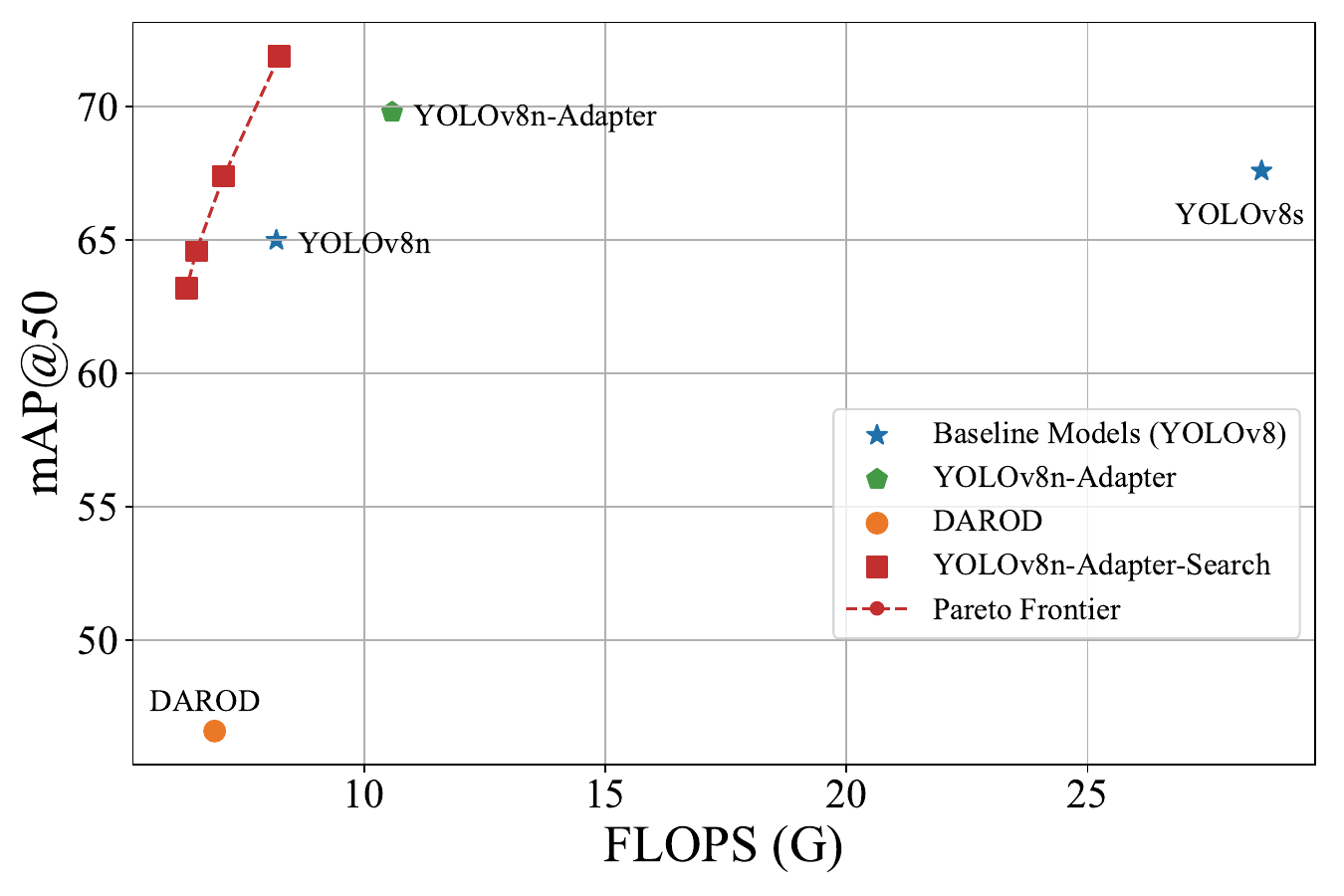}
    \end{minipage}
    \begin{minipage}{0.32\linewidth}
        \centering
        \includegraphics[width=\linewidth]{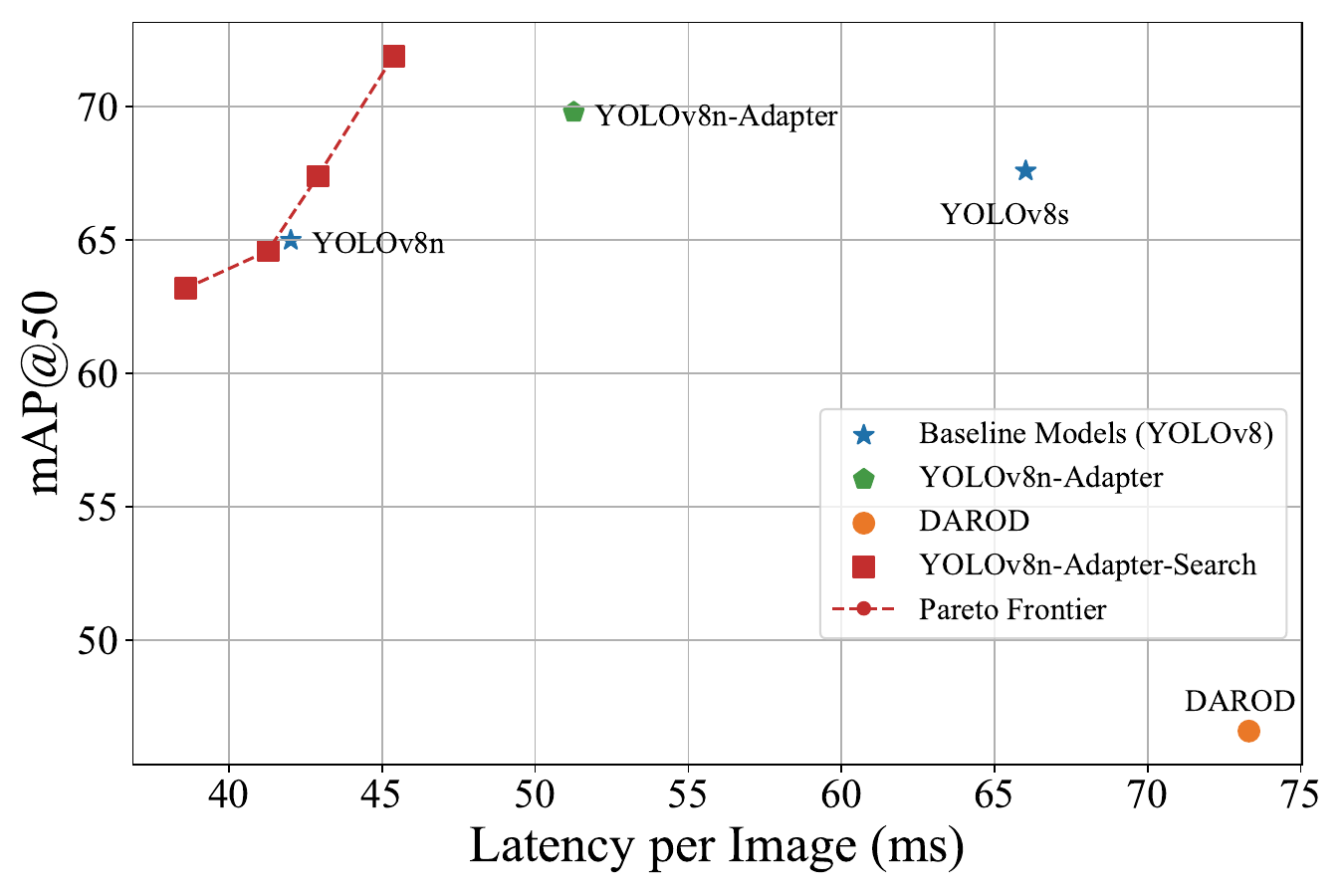}
    \end{minipage}
    \caption{\textbf{Model parameters, GFLOPs, latency \textit{vs.} accuracy on RADDet dataset.} Our method achieves the best accuracy-efficiency trade-offs.}
    \label{fig:mAP vs param&infer&flops}
    \vspace{-12pt}
\end{figure*}

\subsection{Search Results}
Table~\ref{searchResultTable} compares the results of \lzw{YOLOv8} variants and the best subnet on the RADDet and CARRADA datasets.
For YOLOv8 models without the Adapter, we use the heatmap representation of RD maps as the input, \lzw{which provides better performance than the grayscale representation.}
The results indicate that the Adapter branch introduces only \lzw{few additional model} parameters while significantly enhancing model accuracy, \zwc{improving mAP@50-95 from 42.2 to 45.1 on the RADDet dataset and from 28.0 to 29.6 on the CARRADA dataset.}
\lzw{With one-shot NAS, the best subnet, YOLOv8n-Adapter-Search, can further improve} \zwc{mAP@50-95 to 45.4 and 30.3 on RADDet and CARRADA, respectively.}
\zwc{Moreover, as illustrated in Fig.~\ref{fig:mAP vs param&infer&flops}, the Pareto frontier indicates the effectiveness of our search approach in balancing accuracy and inference efficiency, which is particularly crucial for resource-constrained scenarios.} 
Notably, on the CARRADA dataset, deeper YOLOv8 models perform worse due to the limited size of the training set (4319 samples), leading to overfitting.

\begin{table}[t!]
\centering
\caption{\textbf{Generalization performance of the Adapter.} Adapter consistently improves mAP metrics across various YOLO models.}
\vspace{-4pt}
\label{Generalization capability}
\resizebox{0.7\textwidth}{!}{
\begin{tabular}{l|c|c|cc|cc}
\toprule
\textbf{Dataset} & \textbf{Model} & \textbf{Params (M)} & \textbf{mAP@50} & \textbf{$\Delta$} & \textbf{mAP@50-95} & \textbf{$\Delta$} \\
\midrule
\multirow{6}{*}{\textbf{RADDet}}
& YOLOv8n & 3.01 & 65.0 & - & 42.2 & - \\
& {lightgray}\textbf{YOLOv8n-Adapter} & {lightgray}\textbf{3.24} & {lightgray}\textbf{69.8} & {lightgray}\textbf{4.8} & {lightgray}\textbf{45.1} & {lightgray}\textbf{2.9}\\ \cline{2-7}
& YOLOv9t & 1.97 & 63.5 & - & 41.1 \\
& {lightgray}\textbf{YOLOv9t-Adapter} & {lightgray}\textbf{2.16} & {lightgray}\textbf{65.0} & {lightgray}\textbf{1.5} & {lightgray}\textbf{43.2} & {lightgray}\textbf{2.1} \\ \cline{2-7}
& YOLOv10n & 2.70 & 63.5 & - & 41.3 \\
& {lightgray}\textbf{YOLOv10n-Adapter} & {lightgray}\textbf{2.93} & {lightgray}\textbf{67.2} & {lightgray}\textbf{3.7} & {lightgray}\textbf{44.2} & {lightgray}\textbf{2.9} \\
\midrule
\multirow{6}{*}{\textbf{CARRADA}}
& YOLOv8n & 3.01 & 55.5 & - & 28.0 & - \\
& {lightgray}\textbf{YOLOv8n-Adapter} & {lightgray}\textbf{3.24} & {lightgray}\textbf{56.6} & {lightgray}\textbf{1.1} & {lightgray}\textbf{29.6} & {lightgray}\textbf{1.6}\\ \cline{2-7}
& YOLOv9t & 1.97 & 55.8 & - & 27.1 & - \\
& {lightgray}\textbf{YOLOv9t-Adapter} & {lightgray}\textbf{2.16} & {lightgray}\textbf{55.8} & {lightgray}\textbf{0.0} & {lightgray}\textbf{28.3} & {lightgray}\textbf{1.2}\\ \cline{2-7}
& YOLOv10n & 2.70 & 53.6 & - & 28.0 & - \\
& {lightgray}\textbf{YOLOv10n-Adapter} & {lightgray}\textbf{2.93} & {lightgray}\textbf{55.2} & {lightgray}\textbf{1.6} & {lightgray}\textbf{28.6} & {lightgray}\textbf{0.6}\\
\bottomrule
\end{tabular}
}
\end{table}

\subsection{Model Generalization of the Adapter}
\vspace{-5pt}
To evaluate the model generalization of the Adapter Branch, we integrated it into the various YOLO models. As demonstrated in Table~\ref{Generalization capability}, the results reveal that the Adapter achieves significant performance improvements across all baseline models while introducing few additional parameters (<0.23M). 
Notably, the YOLOv8n exhibits the best performance improvement and outperforms other models with the Adapter.
Thus, we choose YOLOv8n as our baseline.

\begin{table}[t!]
\centering
\begin{minipage}{0.48\linewidth} 
\centering
\caption{\textbf{Ablation studies on input representations for RD Radar images.} The best results are highlighted with a gray background.}
\vspace{-4pt}
\resizebox{\columnwidth}{!}{ 
\begin{tabular}{l|c|c|c|c}
\toprule
\textbf{Dataset} & \textbf{Model} & \textbf{Backbone Input} & \textbf{Adapter Input} & \textbf{mAP@50-95} \\
\midrule
\multirow{6}{*}{\textbf{RADDet}} 
& \multirow{2}{*}{YOLOv8n} & heat & -- & 42.2 \\
& & gray & -- & 34.4 \\
\cmidrule{2-5}
& \multirow{4}{*}{YOLOv8n-Adapter} & {lightgray}\textbf{heat} & {lightgray}\textbf{gray} & {lightgray}\textbf{45.1} \\
& & gray & heat & 29.4 \\
& & heat & heat & 43.2 \\
& & gray & gray & 42.1 \\
\midrule
\multirow{6}{*}{\textbf{CARRADA}} 
& \multirow{2}{*}{YOLOv8n} & heat & -- & 28.0 \\
& & gray & -- & 27.2 \\
\cmidrule{2-5}
& \multirow{4}{*}{YOLOv8n-Adapter} & {lightgray}\textbf{heat} & {lightgray}\textbf{gray} & {lightgray}\textbf{29.6} \\
& & gray & heat & 14.4 \\
& & heat & heat & 28.9 \\
& & gray & gray & 27.9 \\
\bottomrule
\end{tabular}}
\label{tab:input_methods_comparison}
\end{minipage}
\hfill 
\begin{minipage}{0.48\linewidth} 
\centering
\caption{\textbf{Ablation studies for Primary-Auxiliary Fusion Module.} The best results are shown in bold, and the second-best results are underlined.}
\resizebox{\columnwidth}{!}{
\begin{tabular}{l|l|c c c}
\toprule
\textbf{Dataset} & \textbf{Fusion Method} & \textbf{mAP@50} & \textbf{mAP@70} & \textbf{mAP@50-95} \\
\midrule
\multirow{3}{*}{\textbf{RADDet}} 
 & Fusion Option 2  & \underline{68.9} &52.1& 42.7 \\
 & Fusion Option 3  & 67.8&\textbf{54.9} & \underline{43.6} \\
\cmidrule(lr){2-5}
 & Fusion Option 1     & \textbf{69.8}&\underline{54.4} & \textbf{45.1} \\
\midrule
\multirow{3}{*}{\textbf{CARRADA}} 
 & Fusion Option 2  & \textbf{57.8}&34.5 & \underline{29.4} \\
 & Fusion Option 3  & 56.0&\underline{35.0} & 29.1 \\
\cmidrule(lr){2-5}
 & Fusion Option 1     & \underline{56.6}&\textbf{37.4} & \textbf{29.6} \\
\bottomrule
\end{tabular}}
\label{ablation for fusion}
\end{minipage}
\vspace{-12pt}
\end{table}

\vspace{-4pt}
\subsection{Ablation Studies}\label{Ablation}
\noindent\textbf{Different Representations of Input.}
We conducted ablation studies on different input representations for radar RD maps, as shown in Table~\ref{tab:input_methods_comparison}. For YOLOv8n without the adapter, using heatmaps as input outperformed grayscale images, demonstrating that heatmaps provide richer information. When the Adapter is introduced, the best performance is achieved by feeding heatmaps to the backbone and grayscale images to the Adapter branch. Conversely, reversing this configuration significantly degrades accuracy, highlighting the complementary roles of the two branches: the backbone extracts high-level features, while the Adapter enriches low-level details. Using identical input representations for both branches leads to inferior performance due to feature redundancy. 
These findings emphasize the importance of tailored inputs to maximize detection performance.

\noindent\textbf{Effectiveness of Primary-Auxiliary Fusion Module.}
To examine the effectiveness of the feature fusion module designed for integrating primary and auxiliary feature maps within the Exchanger Modules, \lzw{we present the results of two simplified fusion methods (fusion Option 2 and Option 3)}, as shown in Table \ref{ablation for fusion}. 
The results show that the feature fusion guided by the proposed channel-level weights of the auxiliary feature map \lzw{achieves the best results}. This is consistent with the search results presented in Section 4.3, where Option 1 predominates among the architecture choices for feature fusion in the optimal subnets.

\begin{table*}[t!]
\centering
\caption{
\textbf{Ablation studies on main components of the Adapter.}
}
\begin{adjustbox}{width=\textwidth}
\begin{tabular}{l |c| c|c|c |c| cc| cc}
\toprule
\textbf{Dataset} & \textbf{Model} & \textbf{Stem} & \textbf{Exchanger Mode 1} & \textbf{Exchanger Mode 2} & \textbf{Params(M)} & \textbf{mAP@50} & \textbf{$\Delta$} & \textbf{mAP@50-95} & \textbf{$\Delta$} \\
\midrule
\multirow{4}{*}{\textbf{RADDet}}
& YOLOv8n &  & & & 3.01 & 65.0 & - & 42.2 & - \\
& Variant-1 & \checkmark & & & 3.10 & 63.6 & -1.4 & 43.6 & 1.4 \\
& Variant-2 & \checkmark & \checkmark & & 3.18 & 67.0 & 2.0 & 44.1 & 1.9 \\
& YOLOv8n-Adapter & \checkmark & \checkmark & \checkmark & 3.24 & 69.8 & 4.8 & 45.1 & 2.9 \\
\midrule
\multirow{4}{*}{\textbf{CARRADA}}
& YOLOv8n &  & & & 3.01 & 55.5 & - & 28.0 & - \\
& Variant-1 & \checkmark & & & 3.10 & 55.6 & 0.1 & 28.5 & 0.5 \\
& Variant-2 & \checkmark & \checkmark & & 3.18 & 55.2 & -0.3 & 28.7 & 0.7 \\
& YOLOv8n-Adapter & \checkmark & \checkmark & \checkmark & 3.24 & 56.6 & 1.1 & 29.6 & 1.6 \\
\bottomrule
\end{tabular}
\end{adjustbox}
\label{component ablation for adapter}
\vspace{-12pt}
\end{table*}

\noindent\textbf{Main Components of the Adapter.}
To investigate the impact of key modules within the Adapter branch, we progressively integrated these \lzw{designs} into the YOLOv8n baseline, as shown in Table \ref{component ablation for adapter}. 
\zwc{Variant-1 employs a Stem module for initial feature extraction from grayscale images, followed by element-wise addition into the Backbone. Variant-2 further incorporates four Exchanger Modules of Mode 1 and only injects grayscale features from the Adapter branch into the backbone branch, unlike YOLOv8n-Adapter which has a bidirectional feature interaction process.}
Variant-1 shows limited performance improvements, indicating that simple feature summation is insufficient for effectively \lzw{fusing the heatmap features with grayscale features.}
Next, we introduce the Exchanger Module with Mode 1, \lzw{and the results demonstrate that}
the fusion and attention mechanisms within the Exchanger effectively integrate grayscale features into the backbone branch. 
Finally, adding the Exchanger Module with mode 2 further improves detection performance, suggesting that the grayscale image features fused with advanced heatmap features from the backbone can inject richer information back into the backbone branch, forming a positive closed loop. 
In summary, these results confirm that each component within the Adapter is essential.
\section{Conclusion}
\vspace{-4pt}
This paper presents an efficient object detection model for RD radar maps with heatmaps and grayscale multi-representation.
Specifically, we design an Adapter branch, a Primary-Auxiliary Fusion module, and an Exchanger Module to effectively extract and fuse multi-representation features, respectively.
Additionally, we construct a weight-sharing supernet of the proposed model and employ a One-Shot NAS to obtain better accuracy and efficiency trade-off.
Experimental results on RADDet and CARRADA datasets demonstrate that our model achieves new state-of-the-art performance and obtains the best accuracy and efficiency trade-off.
\vspace{-2pt}

\bibliographystyle{splncs04}
\bibliography{references}

\end{document}